# *Learning to See*: You Are What You See

Memo Akten, Rebecca Fiebrink and Mick Grierson

**Memo Akten**
Artist, Researcher
Department of Computing
Goldsmiths University
 of London
London SE14 6NW
U.K.
m.akten@gold.ac.uk

**Rebecca Fiebrink**
Senior Lecturer
Department of Computing
Goldsmiths University
 of London
London SE14 6NW
U.K.
r.fiebrink@gold.ac.uk

**Mick Grierson**
Professor, Research Leader
Creative Computing Institute
University of the Arts London
45–65 Peckham Road
London SE5 8UF
U.K.
m.grierson@arts.ac.uk

**ABSTRACT**

The authors present a visual instrument developed as part of the creation of the artwork *Learning to See*. The artwork explores bias in artificial neural networks and provides mechanisms for the manipulation of specifically trained–for real-world representations. The exploration of these representations acts as a metaphor for the process of developing a visual understanding and/or visual vocabulary of the world. These representations can be explored and manipulated in real time, and have been produced in such a way so as to reflect specific creative perspectives that call into question the relationship between how both artificial neural networks and humans may construct meaning.

**Summary**

The visual instrument has been used to create a specific work that consists of a real-time, interactive system based on an artificial neural network making predictions on a live camera input. The work can be presented as both an interactive installation and a number of video works made using the same system. We use a novel training system and custom software, facilitating playful interaction and continuous, meaningful human control [1] in ways that offer a range of potential creative affordances not available in any other system. We present examples of the work realized through the instrument, and discuss how some of the various creative affordances operate for the potential development of a range of similar works.

The instrument aspect is part of ongoing research into enabling continuous, meaningful human control of generative deep neural networks, specifically for creative expression. As opposed to extensively developing one single instrument, we prefer to investigate many different methods and develop each method enough to demonstrate its potential to the extent that it can ideally provide a foundation for future research. We believe that this may potentially have more impact and encourage more research in this particular field of continuous, meaningful human control of deep neural networks for creative expression.

Our instrument can be described as an image-based artificial neural network trained on a number of different custom datasets, making predictions on live camera input in real time. Our system processes the live camera image and reconstructs a new image that resembles the input in composition and overall shape and structure, but is of a particular nature and aesthetic as determined by the different datasets, e.g. the ocean, flames, clouds, flowers, etc. We also use the same system to create and present a video artwork (this can be viewed in the accompanying video at https://vimeo.com/332185554 at 00:00–03:02). Figures 1 and 2 show example frames.

**Motivation: Bias**

We have two primary motivations behind this work. One is a conceptual motivation related to bias. The work exposes and amplifies the learned bias in artificial neural networks, demonstrating how critical the training data is to the predictions that the model will make. This is particularly obvious when different models using the same architecture (but trained on different datasets) produce radically different outputs for the same input. Metaphorically speaking, the training data determines the full life experience of the network and ultimately shapes its worldview. When the trained network looks out into the world via the camera, it can only see what it already knows.

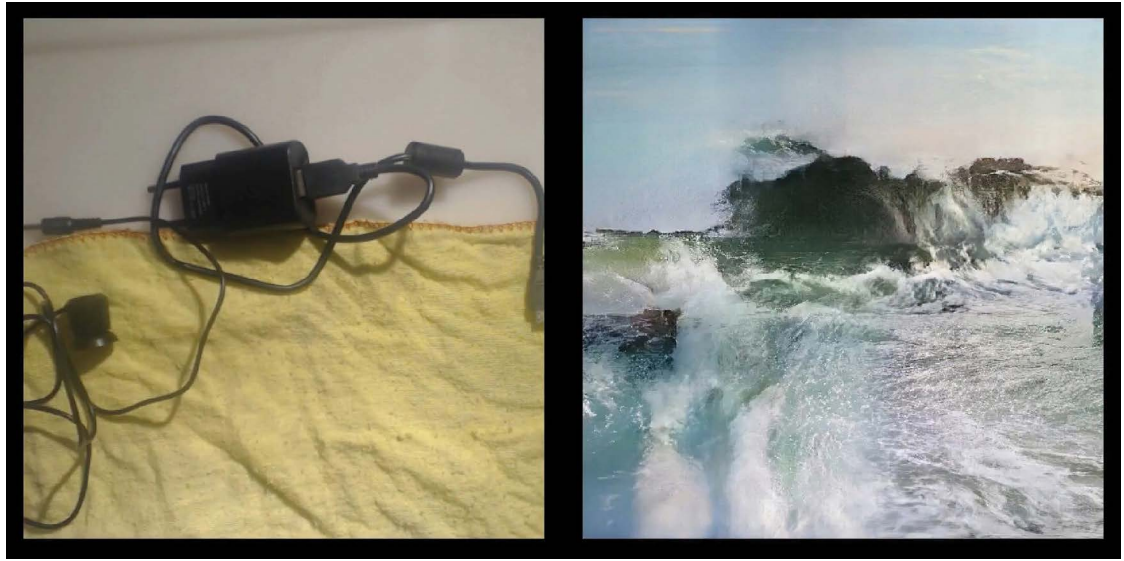

Fig. 1. A frame from *Learning to See* using a model trained on ocean waves. Left: live image from camera. Right: output from the software. (© Memo Akten, 2017)

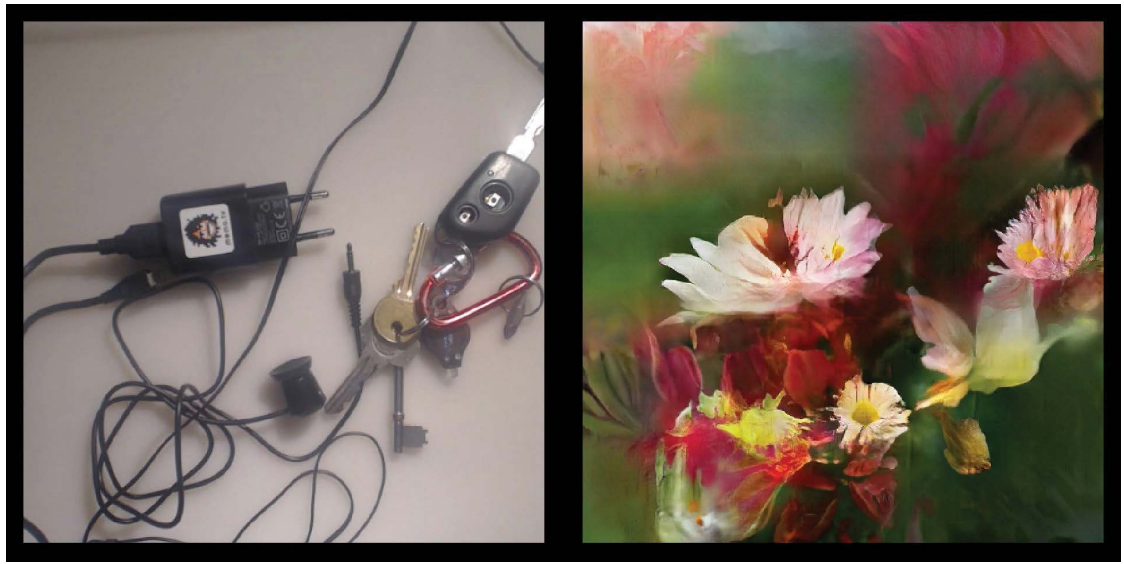

Fig. 2. A frame from *Learning to See* using a model trained on flowers. Left: live image from camera. Right: output from the software.

Thus, at a higher level, the motivation behind the work is to use these convolutional artificial neural networks [2]—very loosely inspired by our own visual cortex [3]—as a reflection on our own self-affirming cognitive biases. This is particularly inspired by models of predictive processing and cognition [4], and more specifically, the idea that the "reality" we perceive in our minds is not a mirror image of the outside world, but a reconstruction based on and limited by our physiology, expectations and prior beliefs [5].

**Motivation: Visual Instruments**

The second motivation behind the work is more practical and is driven by the desire to create visual instruments, real-time, interactive systems for creative expression, with continuous, meaningful human control.

The concept of visual instruments can be traced back to Louis-Bertrand Castel's ocular harpsichord [6] and similar mechanical visual instruments from the eighteenth and nineteenth centuries. Early twentieth-century abstract animators and filmmakers such as Norman McLaren and Oskar Fischinger, analog computer animation pioneer John Whitney Sr. [7] and analog video synthesizers such as the Rutt-Etra are also key influences. The emergence of digital technologies allowed artists such as Myron Krueger, David Rokeby, Ed Tannenbaum, Rebecca Allen, Scott Snibbe, Golan Levin and Camille Utterback, to name a few, to work in this subfield of interactive media art, exploring new modes of expression.

With the current emergence of deep learning (DL), we are able to build hierarchies of representations and extract meaningful information from vast amounts of high-dimensional raw data. Particularly with deep generative models, we believe there is a huge potential in exploring DL in the context of visual instruments.

However, most current DL research is being carried out by specialized researchers whose priorities are not necessarily aligned in this way. Similarly, while many artists—such as Anna Ridler, Mario Klingemann, Helena Sarin and Sofia Crespo—are producing incredible visual works with DL, they too are not focusing on the potential of DL systems as a real-time human-machine collaborative tool or instrument.

We design our system to enable users to create what can be thought of as animated content, in a real-time, interactive manner.

This real-time interactivity and meaningful control comes in two forms. The first is a playful interaction, directly manipulating hands, body or objects in front of a camera. This allows a form of what we call digital puppetry. With such an immediate feedback loop between user actions and output, the user is able to experiment, improvise and perform the visual content. No laborious keyframe placements, no animation curve adjustments, no rendering are necessary; everything is performed (and captured) live, analogous to playing a musical instrument (as can be seen in Figs 1 and 2 and the accompanying video at https://vimeo.com/332185554).

The second form of real-time interactivity comes through a number of parameters (which can be controlled via a MIDI controller, or a graphical user interface [GUI]) that influence how the output image is reconstructed. There are many such parameters that can be adjusted, but the most useful ones we found are those that affect the brightness (or levels) of the output image. But instead of just darkening or brightening the output image, we can guide the network to construct an image using the most appropriate features to produce the desired levels.

For example, adjusting the image to be dark doesn't just darken the output image. It encourages the network to use features that are naturally darker to produce a dark image. We observe similar behavior when brightening the image, inverting it or adjusting the levels. This can be seen quite clearly in Fig. 3 and the accompanying video (https://vimeo.com/332185554) during the segment with the model trained on thousands of images from the Hubble Space Telescope (time stamp 03:00–04:00). At the beginning of that segment, the image is made very dark by lowering the *norm_high* parameter in the GUI. *Norm_low* and *norm_high* are integer values in the range 0 to 255 and control the input image levels. While *norm_high* is low, the input image is dark, so the network uses dark features—in this case distant galaxies—to construct the output image. For areas of relative brightness, the network automatically chooses to use larger, brighter galaxies. As the *norm_high* parameter is increased, the network uses brighter features; it gradually brings in nebulous gas clouds to construct the output image in an increasingly brighter fashion.

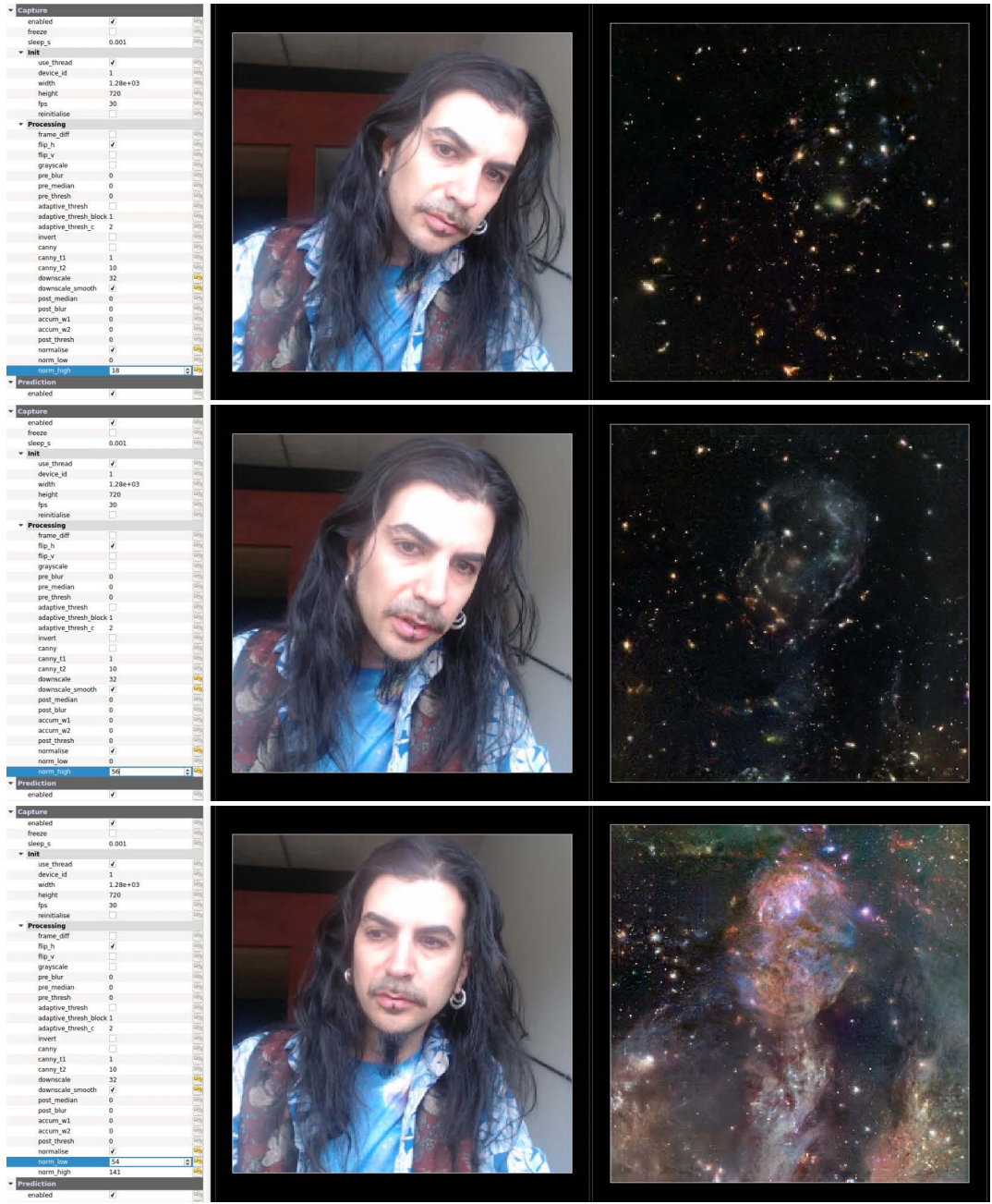

**Fig. 3.** Darkening the image (via the *norm_high* parameter) forces the network to use dark features to construct the output, in this case, distant galaxies. For brighter areas, the network chooses slightly brighter features such as larger and brighter galaxies. As we brighten the image, the network starts using brighter features, such as nebulae.

### Network Architecture and Training

The network architecture we use is based on pix2pix [8], for paired image-to-image translation. Given corresponding image pairs consisting of an input image and a target image, the network learns a transformation from input to target. We have built on this architecture to i) produce higher-resolution images, ii) encourage the network to generalize in the way we require and iii) allow scope for real-time manipulation of parameters to influence the output in a meaningful and expressive manner.

The key difference of our system is in the way we train it. The traditional pix2pix requires corresponding (input, target) image pairs. With our method, we only provide target images. We add the necessary transformations to preprocess the target to the computation graph, so that the input images can be computed on the fly during training. We also parameterize these preprocessing transformations to allow data augmentation during training via random variations. In other words, during training, every time the training loop requests a mini-batch, we load a random selection of target images. These are fed into the preprocessing graph where first, a random crop is taken from each image (a further data augmentation). These regions are then processed via the preprocessing graph with random variations on the parameters to produce a unique mini-batch of input images. Finally, these dynamically created image pairs are used for training (for that mini-batch only). Training in this manner ensures that the network will never see the same image pair twice, but will constantly see different crops and variations, and this encourages the network to generalize, as opposed to memorize, the data. We train for 500,000 iterations with a batch size of 4.

Below we list the operators with the random variation of parameters we found to produce the most desirable results (after an extensive random search of the parameter space):

- scale target image to a random size between 100% and 130% of the desired image size
- take random crop of the desired size
- depending on the nature of the dataset, randomly flip horizontally and/or vertically
- convert to grayscale
- downscale 24x with area filtering, upscale back to original size with cubic filtering
- apply random brightness of up to ±20%
- apply random contrast of up to ±15%

Once trained in this way, the network effectively learns to reconstruct realistic-looking images from what are very blurry, abstract, gray input images. During inference, we use the same preprocessing graph to preprocess the camera feed into similarly blurry, abstract, gray images, and the model is able to reconstruct realistic-looking images using the features that it has learnt from the dataset (Fig. 4).

**Difference with Style Transfer**

At first glance, this may appear to be similar to neural style transfer [9], in which the "style" of one image (e.g. a painting by Monet) is transferred onto another image, which provides the "content" (e.g. a photograph). However, our method has a number of features that make it more suitable for our needs.

The original style transfer method does not produce a predictive model, but instead produces an output image via an optimization process. Hence it cannot run in real time. There are many newer methods that do produce predictive models, and even use more than one reference style image [10]. However, the number of reference images is generally quite small (in order of tens), and the model does not combine its knowledge across the whole set. Instead the user has to choose which reference image to use or provide manual mixing weights. This can be very useful in some cases; however, our needs and motivations are very different.

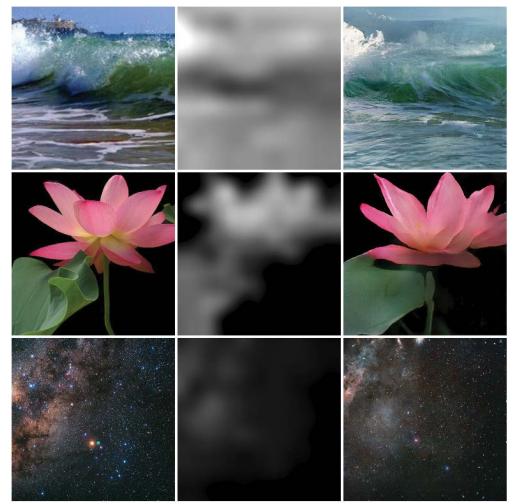

**Fig. 4. Example images generated during training. Left column: random crop from the target image, i.e. ground truth. Center column: input image generated on the fly via real-time processing of the target image crop with randomized parameters. Right column: prediction from the model. Each row is a separate model trained on 5,000–10,000 images of ocean waves, flowers and Hubble images respectively.**

With our system, we can train on large datasets (e.g. 10,000+) of images and learn much richer representations and detail from across the entire dataset. We might even go as far as to claim that the model contains knowledge of the structure of elements related to the contents of the dataset. This can be observed in how the network has learned to generate "foam" around the "rocks" in Fig. 1.

**Conclusion**

The two themes underlying this work are very significant to us. We believe DL to have huge potential in many fields, including acting as a collaborative partner to aid in the production of creative outputs, specifically in a real-time, continuous manner, analogous to playing a musical instrument. We also believe in the magnitude of understanding how vital the distribution of the training data is to the predictions that the network makes. Furthermore, in an increasingly polarized and divided society, we believe in the significance of trying to be sensitive to the idea that we ourselves may be as biased as one of these artificial neural networks, seeing the world through the filter of our own life experiences.